\documentclass[letterpaper]{article} 
\usepackage{aaai25}  
\usepackage{times}  
\usepackage{helvet}  
\usepackage{courier}  
\usepackage[hyphens]{url}  
\usepackage{graphicx} 
\usepackage{natbib}  
\usepackage{caption} 
\usepackage{algorithm}
\usepackage{algorithmic}

\usepackage{newfloat}
\usepackage{listings}
\DeclareCaptionStyle{ruled}{labelfont=normalfont,labelsep=colon,strut=off} 
\floatstyle{ruled}
\newfloat{listing}{tb}{lst}{}
\floatname{listing}{Listing}
\usepackage{multirow} 
\usepackage{amsmath}
\usepackage{booktabs}
\usepackage{amssymb}

\title{Unlearning Concepts in Diffusion Model via \\ Concept Domain Correction and Concept Preserving Gradient}
\author{
    Yongliang Wu\textsuperscript{\rm 1}\equalcontrib, Shiji Zhou\textsuperscript{\rm 2}\equalcontrib\thanks{Corresponding author}, Mingzhuo Yang\textsuperscript{\rm 3}, Lianzhe Wang\textsuperscript{\rm 2}, Heng Chang\textsuperscript{\rm 2}, \\ Wenbo Zhu\textsuperscript{\rm 4}, Xinting Hu\textsuperscript{\rm 5}, Xiao Zhou\textsuperscript{\rm 2}, Xu Yang\textsuperscript{\rm 1}\\
}
\affiliations{
    \textsuperscript{\rm 1}Southeast University\\
    \textsuperscript{\rm 2}Tsinghua University \\
    \textsuperscript{\rm 3}The Chinese University of Hong Kong, Shenzhen \\
     \textsuperscript{\rm 4}University of California, Berkeley \\
     \textsuperscript{\rm 5}Nanyang Technological University\\

    yongliangwu@seu.edu.cn, zhoushiji00@gmail.com
}

\usepackage{bibentry}

\begin{document}

\maketitle

\begin{abstract}
Text-to-image diffusion models have achieved remarkable success in generating photorealistic images. However, the inclusion of sensitive information during pre-training poses significant risks. Machine Unlearning (MU) offers a promising solution to eliminate sensitive concepts from these models. Despite its potential, existing MU methods face two main challenges: 1) limited generalization, where concept erasure is effective only within the unlearned set, failing to prevent sensitive concept generation from out-of-set prompts; and 2) utility degradation, where removing target concepts significantly impacts the model's overall performance. To address these issues, we propose a novel concept domain correction framework named \textbf{DoCo} (\textbf{Do}main \textbf{Co}rrection). By aligning the output domains of sensitive and anchor concepts through adversarial training, our approach ensures comprehensive unlearning of target concepts. Additionally, we introduce a concept-preserving gradient surgery technique that mitigates conflicting gradient components, thereby preserving the model's utility while unlearning specific concepts. Extensive experiments across various instances, styles, and offensive concepts demonstrate the effectiveness of our method in unlearning targeted concepts with minimal impact on related concepts, outperforming previous approaches even for out-of-distribution prompts.

\end{abstract}

\begin{links}
    \link{Code}{https://github.com/yongliang-wu/DoCo}
\end{links}

\section{Introduction}
The development of diffusion models for text-to-image synthesis has progressed rapidly, showcasing a remarkable ability to generate photorealistic images~\cite{liu2024countering}. This has led to the creation of applications such as Stable Diffusion~\cite{rombach2022high} and Midjourney~\cite{borji2022generated}. However, the pre-training datasets often include copyrighted materials, personal photos~\cite{carlini2023extracting}, and unsafe information~\cite{schramowski2023safe}, resulting in generated images that pose copyright, privacy, and safety concerns. This situation raises significant risks of sensitive data leakage~\cite{wu2022membership}, directly conflicting with the growing legislative emphasis on the ``right to be forgotten''~\cite{rosen2011right}. 

\begin{figure*}[t]
    \centering
    \includegraphics[width=\linewidth]{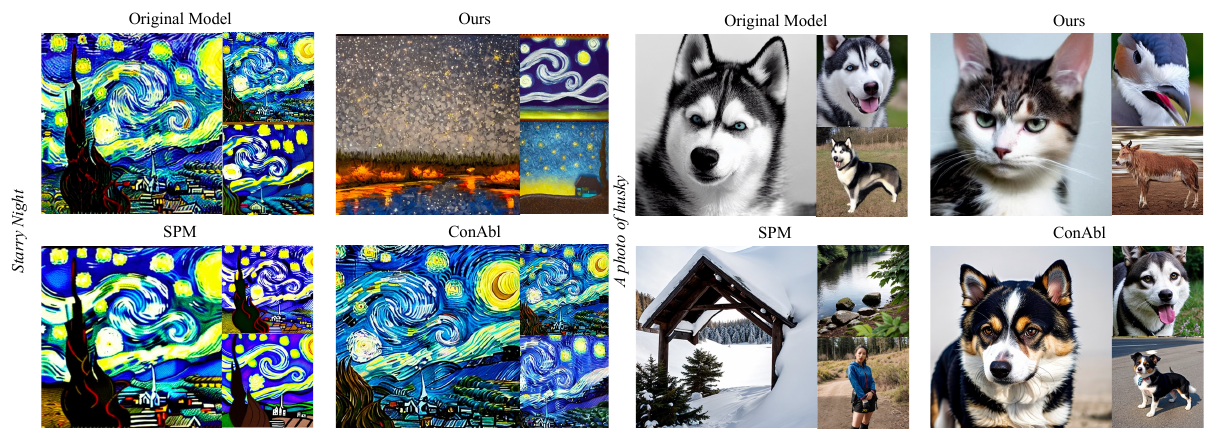}
    \caption{When transferred to strongly related out-of-distribution prompts, previous methods fail to unlearn successfully, whereas our method achieves this generalization. Left: unlearning ``Van Gogh''. Right: unlearning ``Dog''.}
    \label{fig:novelty}
\end{figure*}

A naive approach is to filter out sensitive information and then retrain the entire model from scratch~\cite{rombach2022high,rao2023responsible,kim2025safeguard}. However, given that these generative models are typically trained on large-scale datasets~\cite{schuhmann2022laion}, this method can be prohibitively costly. Another strategy involves using a Safety Checker to detect the presence of inappropriate content in generated images~\cite{rombach2022high,liu2024jailbreak,wang2024aeiou,gao2024meta,yoon2024safree}. However, this approach relies heavily on the performance of the detector and is limited by the biases.

In addressing these challenges, Machine Unlearning (MU)~\cite{xu2023survey,gao2024practical,gao2024semantic,zhou2024limitations} has emerged as a potentially promising solution. Fundamentally, MU is a method that enables models to unlearn the memory of sensitive information contained within their training datasets, thereby preventing the generation of images involving sensitive concepts through crafted prompts. Recently, a series of studies have been introduced that focus on the unlearning concepts in diffusion models~\cite{gandikota2023erasing,zhang2023forget,gandikota2024unified,kumari2023ablating,lyu2023one,zhang2024endowing,guo2024multi,liu2023trapdoor,zhang2024adversarial,truong2024attacks}. For instance, ESD~\cite{gandikota2023erasing} fine-tunes the model to predict in the opposite direction of classifier-free guidance, effectively aligning the targeted concept with that of an empty string. ConAbl~\cite{kumari2023ablating} proposes linking a target concept to a predefined anchor concept by minimizing the L2 distance between the predicted noises for these two concepts. SPM~\cite{lyu2023one} employs a Latent Anchor approach to eliminate concepts and uses a similarity-based retention loss to preferentially weight surrogate concepts, thereby preserving non-target concepts.

However, when applying MU for concept unlearning, two primary issues arise. First, there is the challenge of generalizability: ensuring that a model completely unlearns a target concept with only a limited number of training samples. Second, there is the need for concept preservation: ensuring that while unlearning target concepts, the model's utility on other concepts remains unaffected. Existing methods have not fully resolved these issues. Specifically, for unlearning generalization, some approaches focus on reducing the distance between sample pairs generated from the target and anchor concepts~\cite{kumari2023ablating} or minimize the value and key in the cross-attention layers between target text embeddings and anchor text embeddings~\cite{gandikota2024unified}. These methods are effective only for sampled or related prompts and fail to achieve complete concept elimination. For concept preservation, some methods~\cite{kumari2023ablating, lyu2023one, gandikota2024unified} incorporate regularization strategies that, while attempting to retain other concepts, overlook the inherent contradiction between unlearning and utility preservation, thus failing to maintain overall effectiveness.

To address these challenges, we introduce a novel unlearning framework named \textbf{DoCo} (\textbf{Do}main \textbf{Co}rrection). This framework employs a discriminator to simulate a membership inference attack (MIA)~\cite{shokri2017membership}, distinguishing between the output domains of target and anchor concepts. Through adversarial training against MIA, we align these output domains, thereby aiming for the complete elimination of the target concept in a distributional sense. To the best of our knowledge, we are the first to focus on the generalizability of unlearning concepts in text-to-image diffusion models using adversarial training. We find that DoCo performs well with certain prompts strongly associated with a concept, even if not explicitly mentioned, termed Out-of-Distribution (OOD) prompts. For instance, ``Van Gogh'' and ``Starry Night'' are linked. As shown in Figure~\ref{fig:novelty}, our method successfully unlearns ``Starry Night'' and ``Husky'' without it being in the training set, outperforming other methods.

To tackle the issue of concept preservation, we propose a concept-preserving gradient approach. This approach trims the conflicting parts of the unlearning gradient with utility regularization, ensuring that each optimization iteration does not compromise the model's utility. This allows us to achieve both concept unlearning and preservation simultaneously. Through comprehensive experiments across a wide range of instances, styles, and offensive concepts, we demonstrate that our method effectively unlearns targeted concepts while having a minimal impact on closely related surrounding concepts that should be preserved.

Our contributions are summarized as follows: (1) We formulate a novel framework DoCo for unlearning concepts in diffusion models. By employing a distribution classifier in an adversarial manner, we align the output domains of the target concepts to be forgotten with those of the anchor concepts. This approach enhances the generalizability of the unlearning effect. (2) We propose a concept-preserving gradient surgery method. By trimming the contradictory parts between the unlearning gradient and the retraining gradient, our method aims to optimize the unlearning objectives while minimally affecting the model's performance on other concepts. (3) Our extensive experiments demonstrate that DoCo can successfully unlearn specific concepts with minimal impact on related concepts. Compared to state-of-the-art methods, our approach even generalizes to strongly related out-of-distribution prompts.

\section{Related Work}
Text-to-image diffusion models~\cite{rombach2022high, saharia2022photorealistic, nichol2022glide,zhang2024can,cheng2024mvpaint,yang2025learn,singh2024negative} have recently shown impressive potential in generating high-quality images, typically trained on large-scale, web-crawled datasets such as LAION-5B~\cite{schuhmann2022laion}. However, these uncurated datasets often include harmful or copyrighted content, which can lead the models to generate sensitive or Not Safe For Work (NSFW) images. The research community has initiated various efforts to address this issue, which can generally be classified into three main categories: dataset filtering, model fine-tuning, and post-generation classification.
\subsection{Dataset Filtering}
Dataset filtering involves excluding specific images from the training dataset. For instance, Adobe Firefly uses licensed and public-domain materials to ensure suitability for commercial use~\cite{rao2023responsible}. Similarly, Stable Diffusion v2.0~\cite{rombach2022high} employs an NSFW filter to eliminate inappropriate content from the LAION-5B dataset. While this approach has demonstrated some effectiveness, it presents notable challenges. Firstly, it can impose significant operational burdens, as it may necessitate retraining the model from scratch. Secondly, the accuracy and inherent biases of these filtering systems can lead to unreliable exclusion of sensitive content. Additionally, detecting abstract copyright elements, such as an artist's unique style, remains challenging for detection methods.

\subsection{Model Fine-tuning}
Model fine-tuning methods focus on adjusting the model's weights to prevent the generation of harmful content~\cite{zhang2025generate,zhang2024defensive,park2024direct,ko2024boosting,kim2025race}. ESD~\cite{gandikota2023erasing} aligns the probability distributions of the targeted concept with that of an empty string. However, this technique may lead to a collapse problem due to the lack of explicit control. Forget-Me-Not~\cite{zhang2023forget} employs attention re-steering to identify and modify attention maps associated with specific concepts within the cross-attention layers of the diffusion U-Net architecture. However, both methods lack a mechanism for preserving the integrity of other non-target concepts. UCE~\cite{gandikota2024unified} uses a closed-form solution to adjust cross-attention weights, recalibrating them to induce deliberate changes in the keys and values associated with specific text embeddings for the concepts to be edited, while aiming to limit the impact on a separate set of concepts to be retained. ConAbl~\cite{kumari2023ablating} suggests associating a target concept with a predefined anchor concept by minimizing the L2 distance between the predicted noises for these two concepts, and introduces a regularization loss to preserve the integrity of the anchor concept. However, these approaches do not fully resolve the tension between unlearning and retention objectives, potentially impacting the optimization efficiency of both goals. SPM~\cite{lyu2023one} introduces a novel Latent Anchoring fine-tuning strategy combined with a similarity-based retention loss to differentially weight surrogate concepts that are distant in the latent space. While this approach can alleviate the contradictions between the two objectives, it does not directly resolve their conflicts. Our method directly employs a gradient surgery technique to manipulate the gradients of the two conflicting objective functions. Furthermore, previous approaches heavily depend on the configuration of the training set, such as forming pairs of anchor and target concepts or directly manipulating concepts, which can lead to decreased generalization performance when applied beyond the training scope. In contrast, our method modifies the concept at the distribution level, effectively mitigating this problem.
\subsection{Post-generation Classification}
Post-generation classification methods employ a safety checker~\cite{rombach2022high} to detect images after the generation, determining whether harmful content has been produced. This approach shares a similar drawback with dataset filtering, as it heavily relies on the performance of the classifier, thus impacting the effectiveness of content filtering. Additionally, it struggles to adequately protect against copyright-protected content, which requires specially designed and trained detectors.

\section{Method}
\begin{figure*}[t]
    \centering
    \includegraphics[width=\linewidth]{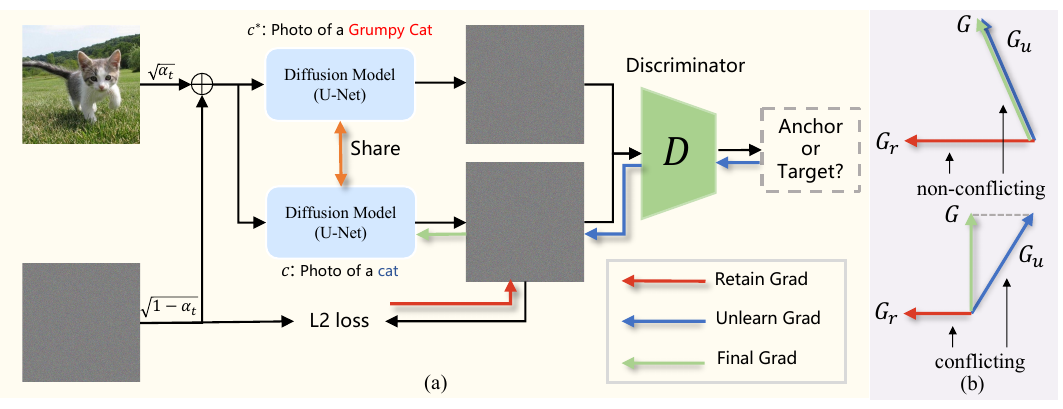}
    \caption{(a) The overall architecture of \textbf{DoCo}, which updates the model parameters through an adversarial training process. This process compels the diffusion model (acting as the generator) to produce denoised results that the discriminator cannot reliably classify as being associated with the target concept, such as ``Grumpy Cat'', or the anchor concept, such as ``Cat''. (b) If the unlearning gradient \(\mathbf{G}_{u}\) does not conflict with the retraining gradient \(\mathbf{G}_{r}\), we update the parameters in the direction of \(\mathbf{G}_{u}\). If \(\mathbf{G}_{u}\) conflicts with \(\mathbf{G}_{r}\), we mitigate the contradictory gradient between them.}
    \label{fig:method}
\end{figure*}

\subsection{Diffusion Models}\label{sec:dm}
Diffusion Models consist of two distinct Markov chains: the forward chain, which incrementally converts the original data into a noisy state, and the reverse chain, which meticulously reconstructs the original data from this noisy state~\cite{song2020denoising,ho2020denoising}. In the diffusion model forward process, noise $\epsilon$ is progressively added to the input image across multiple timesteps $t \in [0,T]$. The noisy image at timestep $t$ can be generated according to the following equation:
\begin{equation}
    x_t = \sqrt{\alpha_t}x_{0} + \sqrt{1 - \alpha_t}\epsilon,
    \label{eq:addnoise}
\end{equation}
 where $x_0$ represents the original clean image and $\alpha_t$ modulates the noise level. 
 
The denoising network $\epsilon_\theta(x_t, c, t)$ is adeptly trained to reverse the noise addition, reconstructing the image at timestep $t-1$ from the noisy image at timestep $t$. This network can also incorporate conditions from other modalities, such as a caption $c$, to guide the denoising process~\cite{rombach2022high}. This process can be formulated as follows:
\begin{equation}
    p_{\theta}(\mathbf{x}_0, \ldots, \mathbf{x}_T) = p(\mathbf{x}_{T}) \prod_{t=1}^T p_{\theta}(\mathbf{x}_{t-1}\vert\mathbf{x}_t),
    \label{eq:reverse}
\end{equation}
where $p(\mathbf{x}_{T})$ is the distribution of the data after adding noise over $T$ timesteps, and $p_{\theta}(\mathbf{x}_{t-1}\vert\mathbf{x}_t)$ represents the conditional probability of recovering the data at the timestep $t-1$ from the noisy data at timestep $t$.

\subsection{Concept Unlearning Formulation}\label{sec:uc}
We define the process of unlearning concepts as the removal of a specific concept \( c^* \) from the pre-trained model \( M_{\text{init}} \). Assuming an unlearning algorithm \( \mathcal{U} \) that eliminates \( c^* \) from \( M_{\text{init}} \), the parameters of \( M_{\text{init}} \) will be altered as a result of this operation. The modified model is denoted as \( \hat{M} \), and the process can be represented by the following formula:
\begin{equation}
    \mathcal{U}(M_{\text{init}}, c^*) = \hat{M}.
\end{equation}

Previous methods~\cite{kumari2023ablating,gandikota2023erasing,lyu2023one} typically align the target concept \( c^* \) explicitly with an anchor concept \( c \) (which could also be an empty string). This can be mathematically expressed as:
\begin{equation}
    \min_{\hat{M}} \text{dis}(\hat{M}(c^*), M_{\text{init}}(c)) = 0.
\end{equation}
Since previous methods optimize the above problem by minimizing the loss between training prompts containing the target concept \( c^* \) and the anchor concept \( c \), this strategy often heavily depends on the design of the training prompt and lacks the ability to generalize beyond the specific prompt template used during training. Ideally, we aim to achieve an alignment of distributions rather than merely minimizing the distance for a specific training prompt template. This unlearning objective can be formulated as:
\begin{equation}\label{eq:unlearn}
    \min_{\hat{M}} \text{dis}(P(\hat{M}(c^*)), P(M_{\text{init}}(c))).
\end{equation}
Simultaneously, we wish to maintain the distribution of the original anchor and other unrelated concepts unchanged. Thus, the following retaining objective should be satisfied:
\begin{equation}\label{eq:retain}
    P(\hat{M}(c)) \sim P(M_{\text{init}}(c)).
\end{equation}
Therefore, Equation~\ref{eq:unlearn} can also be approximated as:
\begin{equation}\label{eq:unlearn_appro}
    \text{dis}(P(\hat{M}(c^*)), P(\hat{M}(c))) \approx 0.
\end{equation}
We assume that an unlearning algorithm should satisfy both Equation~\ref{eq:unlearn} and Equation~\ref{eq:unlearn_appro} simultaneously.

\begin{figure*}[t]
    \centering
    \includegraphics[width=1\linewidth]{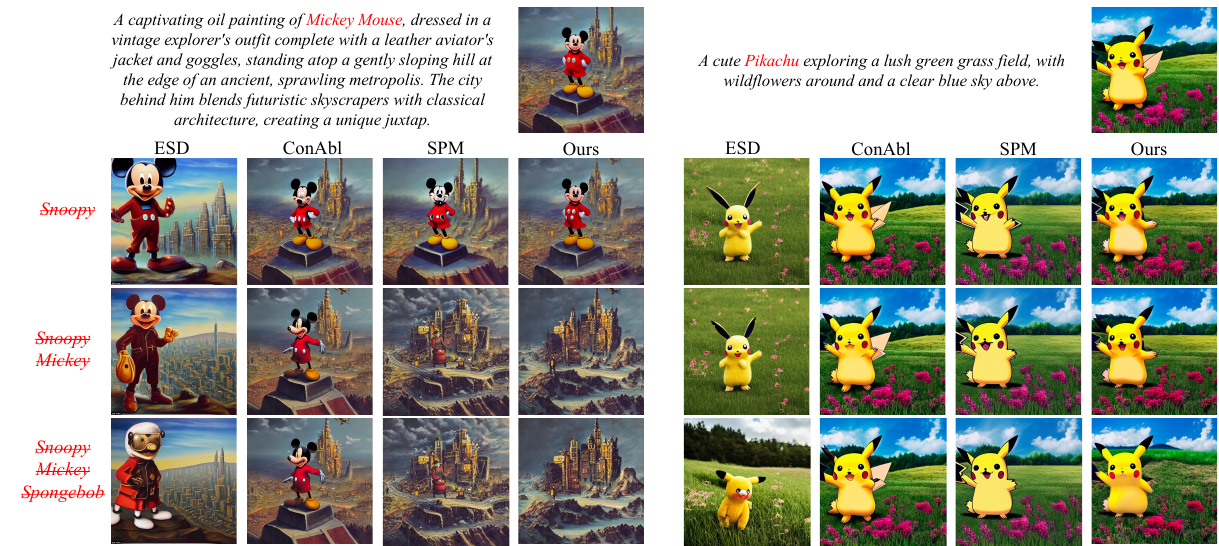}
    \caption{Visualization examples of instance unlearning. The prompts for image generation are displayed at the top. Concepts that have been unlearned are indicated in red text on the left side of the images.}
    \label{fig:instances}
\end{figure*}

\subsection{Concept Domain Correction}\label{sec:cdc}
To address the first optimization objective outlined in Equation~\ref{eq:unlearn}, we propose a Concept Domain Correction framework, based on adversarial training~\cite{goodfellow2020generative}. The goal is to align the target concept domain with the anchor concept domain.

An intuitive approach is to fine-tune the pre-trained model \(M_{\text{init}}\), which acts as a \textit{generator}, and to train a \textit{discriminator} \(D\) alternately until the \textit{discriminator} can no longer distinguish between \(\hat{M}(c)\) and \(\hat{M}(c^*)\). However, the Markov chain nature of Diffusion Models makes direct manipulation of the generated images quite challenging~\cite{rombach2022high}. Therefore, we propose to align the noise distribution directly within the latent space. By adopting this approach, the task of the discriminator shifts from assessing the ``authenticity'' of images~\cite{goodfellow2020generative} to determining the text condition of the currently predicted noise.

As shown in Figure~\ref{fig:method} (a), assume we have a target concept \(c^*\) that we wish to unlearn and an anchor concept \(c\). We can generate samples \(x\) by drawing from the distribution \(p(x|c)\), where \(x\) represents an image generated by the pre-trained diffusion model \(M_{\text{init}}\).Then, the noised samples \(\{x_i\}_{i=0}^{t}\) are obtained by applying the process outlined in Equation~\ref{eq:addnoise}.

The denoising network, represented as \(\epsilon_\theta\), is tasked with estimating the noise at a specific timestep \(t\), conditioned on two distinct conceptual conditions, \(c^*\) and \(c\). After predicting the noise, we perform a denoising step to obtain \(x_{t-1}\). We then construct a discriminator, denoted as \(D\), to simulate a membership inference attack (MIA)~\cite{wu2022membership}, with the goal of distinguishing between the denoised results associated with these two concepts. The parameters of the denoising network are updated to ensure that the discriminator cannot discern which concept the denoised result is conditioned on. We iteratively update the parameters of both the generator and the discriminator, which can be formulated as a min-max objective:

\begin{multline}
\mathcal{L}_{u} = \min_{\epsilon_\theta} \max_D V(\epsilon_\theta, D) = \mathbb{E}_{x \sim p(x|c)}[\log (D(x_{t-1}(c)))] \\
+ \mathbb{E}_{x \sim p(x|c^*)}[\log (1 - D(x_{t-1}(c^*)))].
\end{multline}

\(x_{t-1}(c)\) and \(x_{t-1}(c^*)\) represent the denoised results obtained from the conditions \(c\) and \(c^*\), respectively. 

Theoretically, the framework of adversarial training ensures alignment between the target concept domain and the anchor concept domain~\cite{ganin2016domain}. Since this alignment is achieved at the domain level without directly manipulating the target and anchor concepts, it is anticipated that the generalization capability of concept unlearning will surpass that of prior approaches.

\begin{table*}[t]
\centering
\fontsize{9}{11}\selectfont
\setlength{\tabcolsep}{0.6mm}
\begin{tabular}{c|ccc|ccc|ccc|ccc|ccc|ccc}
    \toprule
    & \multicolumn{3}{c|}{Snoopy} & \multicolumn{3}{c|}{Mickey} & \multicolumn{3}{c|}{Spongebob} & \multicolumn{3}{c|}{Van Gogh} & \multicolumn{3}{c|}{Picasso} & \multicolumn{3}{c}{Rembrandt} \\
    & CS & CA & FID & CS & CA & FID & CS & CA & FID & CS & CA & FID & CS & CA & FID & CS & CA & FID \\
    \midrule
    SD& 74.29 & 93.01 & - & 72.07 & 93.25 & - & 72.80 & 90.38 & - & 74.31 & 100.00 & - & 69.67 & 100.00 & - & 72.46 & 100.00 & - \\
    \midrule
    \multicolumn{10}{c}{\textit{Erasing \textbf{Snoopy}}} & \multicolumn{9}{|c}{\textit{Erasing \textbf{Van Gogh}}} \\
    \midrule
    & CS$\downarrow$ & CA$\downarrow$ & FID$\uparrow$ & CS$\uparrow$ & CA$\uparrow$ & FID$\downarrow$ & CS$\uparrow$ & CA$\uparrow$ & FID$\downarrow$ & CS$\downarrow$ & CA$\downarrow$ & FID$\uparrow$ & CS$\uparrow$ & CA$\uparrow$ & FID$\downarrow$ & CS$\uparrow$ & CA$\uparrow$ & FID$\downarrow$ \\
    \midrule
    ESD & \underline{50.77} & \underline{38.62} & \underline{150.22} & 54.88 & 31.70 & 120.55 & 56.55 & 39.21 & 125.55 & 50.64 & \underline{33.58} & 195.76 & 63.48 & 88.72 & 94.88 & 65.10 & 79.02 & 93.35 \\
    ConAbl & 63.47 & 74.87 & 88.79 & 69.23 & 83.37 & 51.77 & 70.12 & \underline{86.37} & 58.08 & 54.60 & 41.26 & 180.47 & 62.83 & 97.34 & 95.93 & 65.96 & 96.50 & 87.54 \\
    SPM & 55.15 & 40.25 & 110.22 & \underline{71.63} & \textbf{92.12} & \textbf{26.45} & \textbf{72.53} & 81.00 & \underline{31.22} & \underline{51.80} & 42.19 & \underline{198.65} & \textbf{68.96} & \textbf{100.00} & \textbf{35.39} & \underline{70.53} & \underline{97.85} & \underline{56.12} \\
    Ours & \textbf{49.10} & \textbf{35.77} & \textbf{204.37} & \textbf{71.84} & \underline{89.24} & \underline{33.21} & \underline{71.18} & \textbf{87.20} & \textbf{30.43} & \textbf{49.55} & \textbf{31.08} & \textbf{215.76} & \underline{66.21} & \textbf{100.00} & \underline{65.44} & \textbf{71.22} & \textbf{98.13} & \textbf{53.16} \\
    \midrule
    \multicolumn{10}{c}{\textit{Erasing \textbf{Snoopy} and \textbf{Mickey}}} & \multicolumn{9}{|c}{\textit{Erasing \textbf{Picasso}}} \\
    \midrule
    & CS$\downarrow$ & CA$\downarrow$ & FID$\uparrow$ & CS$\downarrow$ & CA$\downarrow$ & FID$\uparrow$ & CS$\uparrow$ & CA$\uparrow$ & FID$\downarrow$ & CS$\uparrow$ & CA$\uparrow$ & FID$\downarrow$ & CS$\downarrow$ & CA$\downarrow$ & FID$\uparrow$ & CS$\uparrow$ & CA$\uparrow$ & FID$\downarrow$ \\
    \midrule
    ESD & \underline{50.30} & 46.05 & \underline{177.82} & \underline{49.54} & \underline{30.15} & \underline{197.80} & 50.25 & 17.75 & 167.28 & 67.65 & 74.41 & 94.43 & 57.45 & \underline{32.38} & 170.59 & 69.00 & 76.12 & 81.24 \\
    ConAbl & 61.64 & 71.25 & 106.92 & 62.29 & 63.87 & 106.27 & 69.78 & 80.37 & 69.00 & 66.70 & 94.23 & 119.26 & 55.45 & 37.06 & 210.29 & 69.85 & 92.65 & 82.06 \\
    SPM & 54.75 & \underline{39.50} & 111.07 & 54.00 & 28.87 & 129.97 & \textbf{71.20} & \underline{80.21} & \underline{37.81} & \underline{73.55} & \underline{95.11} & \underline{43.70} & \underline{49.22} & 38.96 & \underline{269.58} & \underline{71.22} & \underline{97.05} & \underline{53.89} \\
    Ours & \textbf{46.53}& \textbf{38.11} & \textbf{210.88} & \textbf{47.35} & \textbf{27.77} & \textbf{212.06} & \underline{70.88}& \textbf{85.44} & \textbf{34.52}& \textbf{73.87} & \textbf{97.81} & \textbf{42.23} & \textbf{48.91} & \textbf{32.15} & \textbf{303.44} & \textbf{72.30} & \textbf{98.09} & \textbf{41.58} \\
    \midrule
    \multicolumn{10}{c}{\textit{Erasing \textbf{Snoopy}, \textbf{Mickey} and \textbf{Spongebob}}} & \multicolumn{9}{|c}{\textit{Erasing \textbf{Rembrandt}}} \\
    \midrule
    & CS$\downarrow$ & CA$\downarrow$ & FID$\uparrow$ & CS$\downarrow$ & CA$\downarrow$ & FID$\uparrow$ & CS$\downarrow$ & CA$\downarrow$ & FID$\uparrow$ & CS$\uparrow$ & CA$\uparrow$ & FID$\downarrow$ & CS$\uparrow$ & CA$\uparrow$ & FID$\downarrow$ & CS$\downarrow$ & CA$\downarrow$ & FID$\uparrow$ \\
    \midrule
    ESD & \underline{49.50} & 43.25 & \underline{201.10} & \underline{47.50} & \underline{28.95} & \underline{200.94} & \underline{45.27} & \textbf{12.33} & \underline{193.63} & 64.83 & 83.95 & 95.26 & 66.14 & 80.51 & 66.74 & 34.48 & \textbf{30.19} & 220.91 \\
    ConAbl & 61.42 & 69.87 & 107.11 & 62.12 & 63.75 & 106.01 & 68.47 & 78.62 & 61.14 & 65.02 & 94.64 & 101.18 & 65.81 & 98.20 & 62.75 & 53.53 & 39.76 & 133.64 \\
    SPM & 54.61 & \underline{40.50} & 112.33 & 54.12 & 29.75 & 128.80 & 52.36 & \underline{23.12} & 152.16 & \underline{73.13} & \textbf{100.00} & \underline{46.89} & \underline{69.26} & \textbf{100.00} & \textbf{34.26} & \textbf{32.69} & 36.98 & \textbf{275.29} \\
    Ours & \textbf{47.21} & \textbf{35.75} & \textbf{210.54}& \textbf{46.07}& \textbf{27.88} & \textbf{209.69}& \textbf{45.22}& 31.65& \textbf{200.21}& \textbf{73.50} & \textbf{100.00} & \textbf{44.20} & \textbf{69.35} & \textbf{100.00} & \underline{40.50} & \underline{33.05} & \underline{31.05} & \underline{260.20} \\
    \bottomrule
\end{tabular}
\caption{Quantitative results of instance unlearning and artist style unlearning. The top-performing results are highlighted in bold, and the second-best results are underlined.}
\label{tab:merged_results}
\end{table*}

\subsection{Concept Preserving Gradient}\label{sec:gradient}
Ensuring the retention of non-target concepts after a specific concept has been unlearned is a crucial evaluation metric for unlearning methods~\cite{nguyen2022survey}. This forms the second optimization objective, as illustrated in Equation~\ref{eq:retain}. Previous methods focus on utilizing a relearning strategy, which involves adding the standard diffusion loss~\cite{kumari2023ablating} to anchor concepts or creating a preservation set to ensure retention: 
\begin{equation}
\mathcal{L}_{r} = \left\| \epsilon - \epsilon_\theta(x_t, c, t) \right\|_2^2
\end{equation}
Although this approach has proven partially feasible, it can still lead to a conflict between the objectives of unlearning and retaining. We adopt an effective gradient surgery paradigm~\cite{yu2020gradient, zhu2023prompt} to directly mitigate the conflicts between the two objectives. As illustrated in Figure~\ref{fig:method} (b), we formally define the gradients of the unlearning and retaining objectives as \(\mathbf{G}_{u}\) and \(\mathbf{G}_{r}\), respectively. The training process presents two scenarios: (1) If the angle between \(\mathbf{G}_{u}\) and \(\mathbf{G}_{r}\) is less than \(90^\circ\), the optimization for the unlearning loss aligns with the retaining objective, so we align the updated gradient \(\mathbf{G}\) with \(\mathbf{G}_{u}\). (2) If the angle exceeds \(90^\circ\), a conflict arises, and optimizing along \(\mathbf{G}_{u}\) would interfere with non-target concepts. To resolve this, we project \(\mathbf{G}_{u}\) onto the orthogonal direction of \(\mathbf{G}_{r}\), thus avoiding interference with the retaining goal. This concept-preserving gradient approach can be succinctly expressed through the following mathematical formulation:
\begin{equation}
\mathbf{G} = 
\begin{cases}
\mathbf{G}_{u}, & \text{if } \mathbf{G}_{u} \cdot \mathbf{G}_{r} \geq 0 \\
\mathbf{G}_{u} - \lambda \cdot \frac{\mathbf{G}_{u} \cdot \mathbf{G}_{r}}{\|\mathbf{G}_{r}\|^{2}} \mathbf{G}_{r}, & \text{otherwise}.
\end{cases}
\end{equation}
Where $\lambda$ is a hyper-parameter used to control the extent of gradient surgery, which we set to a default value of 1.

Through the approach described above, the conflicting component of the unlearning gradient is eliminated. Consequently, the optimization direction will not impair the utility of the model, allowing us to efficiently unlearn concepts while avoiding damage to other content.

\section{Experiments}
\subsection{Implement Details}
We train the model for 2,000 iterations, using a batch size of 8 and a learning rate of 6e-6. During the initial 1000 iterations, we employ a warm-up strategy that updates only the parameters of the discriminator. For the discriminator, we utilize a PatchGAN architecture~\cite{zhu2017unpaired}, which is composed of 5 convolutional layers. Each layer employs convolutional kernels $4\times4$ with a stride of 2. All experiments are carried out on two A100 GPUs.

\begin{figure*}[t]
    \centering
    \includegraphics[width=1\linewidth]{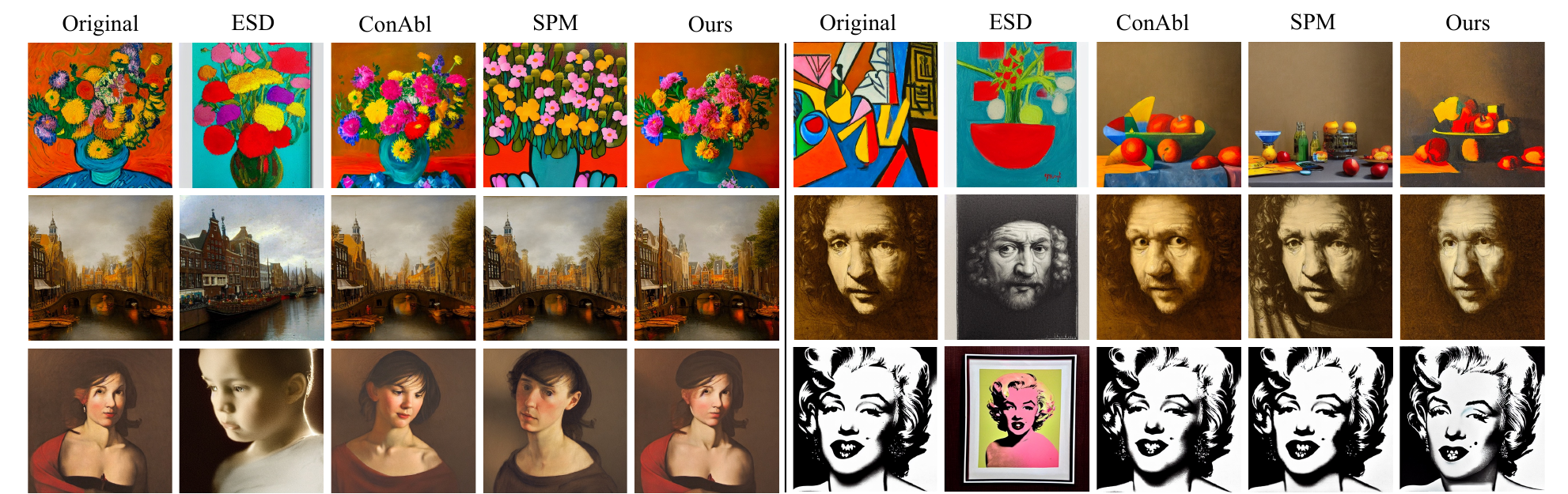}
    \caption{Visualization examples of artistic styles unlearning. Left: Unlearning "Van Gogh". Right: Unlearning "Picasso". The first row represents the forgotten style, while the subsequent rows represent other non-target concepts.}
    \label{fig:style}
\end{figure*}

\subsection{Evaluation metrics}
We utilize the CLIP Score (CS), CLIP Accuracy (CA) and FID as metrics to evaluate the performance on the unlearning of instances and styles~\cite{hessel2021clipscore,kumari2023ablating,heusel2017gans,li2024frame,liu2024envisioning,li2024self}. A concept to be evaluated is incorporated into 80 templates, with each template yielding 10 images. For unlearning celebrities, We use Identity Score Matching (ISM) which compute the cosine distance between the embedding of the generated face and the average face embedding of the entire clean image set~\cite{van2023anti}.

\begin{table}[t]
\centering
\fontsize{9}{11}\selectfont
\setlength{\tabcolsep}{0.4mm}
\begin{tabular}{c|c|c|c}
\toprule
 & Cristiano Ronaldo & LeBron James & Robert Downey Jr\\
\midrule
SD & 0.75 & 0.71 & 0.58 \\
\midrule
\multicolumn{4}{c}{\textit{Erasing \textbf{Cristiano Ronaldo}}} \\
\midrule
 & ISM↑ & ISM↓ & ISM↓ \\
 \midrule
ESD & 0.85 & 0.72 & 0.67 \\
ConAbl & 0.75 & 0.70 & \textbf{0.62} \\
SPM & \textbf{0.89} & 0.77 & 0.64 \\
Ours & \textbf{0.89} & \textbf{0.69} & \textbf{0.62} \\
\midrule
\multicolumn{4}{c}{\textit{Erasing \textbf{LeBron James}}} \\
\midrule
 & ISM↓ & ISM↑ & ISM↓ \\
 \midrule
ESD & 0.74 & 0.90 & 0.65 \\
ConAbl & 0.72 & 0.83 & \textbf{0.57} \\
SPM & 0.77 & 0.88 & 0.61 \\
Ours & \textbf{0.65} & \textbf{0.91} & \textbf{0.57} \\
\bottomrule
\end{tabular}
\caption{Unlearning Celebrities: We selected three common celebrities and unlearned two of them. Our method successfully retains the other celebrities while accurately forgetting the target concept.}
\label{table:celebrities}
\end{table}

\subsection{Main Results}
\paragraph{\textbf{Removal of Instance.}}
We select ``Snoopy'' as the anchor sample and utilize the CLIP text tokenizer's dictionary to identify the most related concepts by cosine similarity: ``Mickey'', ``SpongeBob'', and ``Pikachu''~\cite{kumari2023ablating, lyu2023one}. 

As shown in Table~\ref{tab:merged_results}, our method demonstrates superior performance in unlearning instance concepts, surpassing previous methods. This advantage is maintained when transfer to multi-concept cases. DoCo also exhibits superior performance in retaining non-target concepts, achieving an optimal trade-off.  In contrast, other methods either show significant unlearning, such as ESD, or display inadequate unlearning capabilities while retaining non-target concepts, such as ConAbl and SPM.

We provide some visual results in Figure~\ref{fig:instances}. It can be observed that ESD lacks the ability to retain other non-target concepts during unlearning, as seen in the ``Pikachu'' case. Additionally, ESD struggles with complex prompts, leading to insufficient unlearning, as demonstrated in the ``Mickey'' case. Although ConAbl maintains good retention performance, it fails to achieve effective unlearning. Both SPM and DoCo demonstrate strong unlearning and retention capabilities. However, DoCo excels in preserving aspects such as color and texture. 

\paragraph{\textbf{Removal of Styles.}} 
We selected several representative concepts to verify the performance of unlearning styles, as detailed in Table~\ref{tab:merged_results}. Our approach successfully unlearns specific styles while excelling in preserving non-target artistic styles. Compared to previous works, our method achieves superior results in most cases. We present several qualitative examples in Figure~\ref{fig:style}. We observe that both ESD and SPM exhibit excessive unlearning for ``Van Gogh'', resulting in aesthetically unappealing images. In contrast, our method produces more natural and visually coherent images. ESD consistently underperforms in maintaining other styles across all cases, while our approach excels in retaining performance, especially in preserving ``Caravaggio'' after unlearning ``Van Gogh''.

\paragraph{\textbf{Removal of Celebrities.}} 
Table \ref{table:celebrities} illustrates our method's effectiveness in unlearning celebrities such as Cristiano Ronaldo and LeBron James. Our approach consistently delivers strong results, outperforming other methods. Additionally, it maintains a good retention of other concepts after forgetting specific ones.

\begin{table}[t]
\centering
\fontsize{9}{11}\selectfont
\setlength{\tabcolsep}{0.4mm}
\begin{tabular}{cc|ccc|ccc|ccc}
\toprule
\multirow{2}{*}{L2}  & \multirow{2}{*}{CP} & \multicolumn{3}{c|}{Snoopy} & \multicolumn{3}{c|}{Mickey} & \multicolumn{3}{c}{Spongebob} \\
& & CS $\downarrow$  & CA $\downarrow$ & FID $\uparrow$ & CS $\uparrow$ & CA $\uparrow$ & FID $\downarrow$ & CS $\uparrow$ & CA $\uparrow$ & FID $\downarrow$ \\
\midrule
 \multicolumn{2}{c|}
 {SD} & 74.29 & 93.01 & - & 72.07 & 93.25 & -  & 72.80 & 90.38 & - \\
\midrule
 & & \underline{52.27} & \textbf{31.50} & \underline{146.53} & 66.96 & 77.25 & 107.59  & 69.43 & 81.50 & 85.19 \\
\checkmark & & 56.68 & 60.25 & 187.30 & \underline{70.25} & \textbf{92.00} & \underline{47.54} & \underline{71.16} & \underline{86.75} & \underline{45.55} \\
 \checkmark & \checkmark & \textbf{49.10} & \underline{35.77} & \textbf{204.37} & \textbf{71.84} & \underline{89.24}& \textbf{33.21} & \textbf{71.18} & \textbf{87.20}& \textbf{30.43}\\
\bottomrule
\end{tabular}
\caption{Ablation Study on Assessing the Impact of L2 Loss (L2) vs. Concept Preserving Gradient (CP). It can be observed that without any retain loss, the unlearning process fails to preserve other concepts effectively. On the other hand, directly using L2 loss leads to a degradation in unlearning performance. Our proposed CP method strikes a balance between the two, achieving better results in both unlearning and concept retention.}
\label{tab:ablation}
\end{table}

\subsection{Ablation Study}
To further demonstrate the effectiveness of the concept-preserving gradient, we present the results of employing regularization loss (L2 Loss) as well as a baseline without any retaining operations in Table~\ref{tab:ablation}. It can be observed that without any retaining operations, there is a significant impact on other non-target concepts. The use of regularization loss enhances preservation but negatively affects unlearning performance. In contrast, the concept preserving gradient achieves an optimal trade-off by maintaining superior unlearning performance while ensuring effective retraining.

\section{Conclusion}
This paper aims to address the issues of generalization and utility drop in concept unlearning. We propose a novel concept domain correction framework named \textbf{DoCo} (\textbf{Do}main \textbf{Co}rrection). Firstly, we introduce an approach based on adversarial training for concept domain correction. This method achieves the purpose of generalized unlearning of the target concept by adjusting the output domains of both the target and anchor concepts. Secondly, we propose a concept-preserving gradient method based on gradient surgery. This method eliminates the conflicting parts between the unlearning gradient and the retaining gradient, ensuring that the process of unlearning minimally impacts the model's utility. The results demonstrate the ability of DoCo to achieve more generalized concept unlearning while maintaining the generative performance of other concepts.

\section{Acknowledgments}
This work is supported by the National Science Foundation of China (62206048 and 62401327), the Natural Science Foundation of Jiangsu Province (BK20220819), the Fundamental Research Funds for the Central Universities (2242024k30035), and China Postdoctoral Science Foundation (GZB20230358, 2023M741966 and 2024T170464). The Big Data Computing Center of Southeast University supports this research work. We also want to thank the Yanchuangzhixin company for its GPU support.

\bibliography{aaai25}

\end{document}